\documentclass{article}

\usepackage{PRIMEarxiv}
\usepackage[utf8]{inputenc} 
\usepackage[T1]{fontenc}    
\usepackage{hyperref}       
\usepackage{url}            
\usepackage{booktabs}       
\usepackage{amsfonts}       
\usepackage{nicefrac}       
\usepackage{microtype}      
\usepackage{lipsum}
\usepackage{fancyhdr}       
\usepackage{graphicx}       
\graphicspath{{media/}}     
\usepackage{subfig}
\usepackage[english]{babel}
\usepackage[draft]{minted}
\usepackage{caption} 
\makeatletter
\def\PYG@reset{\let\PYG@it=\relax \let\PYG@bf=\relax%
    \let\PYG@ul=\relax \let\PYG@tc=\relax%
    \let\PYG@bc=\relax \let\PYG@ff=\relax}
\def\PYG@tok#1{\csname PYG@tok@#1\endcsname}
\def\PYG@toks#1+{\ifx\relax#1\empty\else%
    \PYG@tok{#1}\expandafter\PYG@toks\fi}
\def\PYG@do#1{\PYG@bc{\PYG@tc{\PYG@ul{%
    \PYG@it{\PYG@bf{\PYG@ff{#1}}}}}}}
\def\PYG#1#2{\PYG@reset\PYG@toks#1+\relax+\PYG@do{#2}}

\@namedef{PYG@tok@w}{\def\PYG@tc##1{\textcolor[rgb]{0.73,0.73,0.73}{##1}}}
\@namedef{PYG@tok@c}{\let\PYG@it=\textit\def\PYG@tc##1{\textcolor[rgb]{0.24,0.48,0.48}{##1}}}
\@namedef{PYG@tok@cp}{\def\PYG@tc##1{\textcolor[rgb]{0.61,0.40,0.00}{##1}}}
\@namedef{PYG@tok@k}{\let\PYG@bf=\textbf\def\PYG@tc##1{\textcolor[rgb]{0.00,0.50,0.00}{##1}}}
\@namedef{PYG@tok@kp}{\def\PYG@tc##1{\textcolor[rgb]{0.00,0.50,0.00}{##1}}}
\@namedef{PYG@tok@kt}{\def\PYG@tc##1{\textcolor[rgb]{0.69,0.00,0.25}{##1}}}
\@namedef{PYG@tok@o}{\def\PYG@tc##1{\textcolor[rgb]{0.40,0.40,0.40}{##1}}}
\@namedef{PYG@tok@ow}{\let\PYG@bf=\textbf\def\PYG@tc##1{\textcolor[rgb]{0.67,0.13,1.00}{##1}}}
\@namedef{PYG@tok@nb}{\def\PYG@tc##1{\textcolor[rgb]{0.00,0.50,0.00}{##1}}}
\@namedef{PYG@tok@nf}{\def\PYG@tc##1{\textcolor[rgb]{0.00,0.00,1.00}{##1}}}
\@namedef{PYG@tok@nc}{\let\PYG@bf=\textbf\def\PYG@tc##1{\textcolor[rgb]{0.00,0.00,1.00}{##1}}}
\@namedef{PYG@tok@nn}{\let\PYG@bf=\textbf\def\PYG@tc##1{\textcolor[rgb]{0.00,0.00,1.00}{##1}}}
\@namedef{PYG@tok@ne}{\let\PYG@bf=\textbf\def\PYG@tc##1{\textcolor[rgb]{0.80,0.25,0.22}{##1}}}
\@namedef{PYG@tok@nv}{\def\PYG@tc##1{\textcolor[rgb]{0.10,0.09,0.49}{##1}}}
\@namedef{PYG@tok@no}{\def\PYG@tc##1{\textcolor[rgb]{0.53,0.00,0.00}{##1}}}
\@namedef{PYG@tok@nl}{\def\PYG@tc##1{\textcolor[rgb]{0.46,0.46,0.00}{##1}}}
\@namedef{PYG@tok@ni}{\let\PYG@bf=\textbf\def\PYG@tc##1{\textcolor[rgb]{0.44,0.44,0.44}{##1}}}
\@namedef{PYG@tok@na}{\def\PYG@tc##1{\textcolor[rgb]{0.41,0.47,0.13}{##1}}}
\@namedef{PYG@tok@nt}{\let\PYG@bf=\textbf\def\PYG@tc##1{\textcolor[rgb]{0.00,0.50,0.00}{##1}}}
\@namedef{PYG@tok@nd}{\def\PYG@tc##1{\textcolor[rgb]{0.67,0.13,1.00}{##1}}}
\@namedef{PYG@tok@s}{\def\PYG@tc##1{\textcolor[rgb]{0.73,0.13,0.13}{##1}}}
\@namedef{PYG@tok@sd}{\let\PYG@it=\textit\def\PYG@tc##1{\textcolor[rgb]{0.73,0.13,0.13}{##1}}}
\@namedef{PYG@tok@si}{\let\PYG@bf=\textbf\def\PYG@tc##1{\textcolor[rgb]{0.64,0.35,0.47}{##1}}}
\@namedef{PYG@tok@se}{\let\PYG@bf=\textbf\def\PYG@tc##1{\textcolor[rgb]{0.67,0.36,0.12}{##1}}}
\@namedef{PYG@tok@sr}{\def\PYG@tc##1{\textcolor[rgb]{0.64,0.35,0.47}{##1}}}
\@namedef{PYG@tok@ss}{\def\PYG@tc##1{\textcolor[rgb]{0.10,0.09,0.49}{##1}}}
\@namedef{PYG@tok@sx}{\def\PYG@tc##1{\textcolor[rgb]{0.00,0.50,0.00}{##1}}}
\@namedef{PYG@tok@m}{\def\PYG@tc##1{\textcolor[rgb]{0.40,0.40,0.40}{##1}}}
\@namedef{PYG@tok@gh}{\let\PYG@bf=\textbf\def\PYG@tc##1{\textcolor[rgb]{0.00,0.00,0.50}{##1}}}
\@namedef{PYG@tok@gu}{\let\PYG@bf=\textbf\def\PYG@tc##1{\textcolor[rgb]{0.50,0.00,0.50}{##1}}}
\@namedef{PYG@tok@gd}{\def\PYG@tc##1{\textcolor[rgb]{0.63,0.00,0.00}{##1}}}
\@namedef{PYG@tok@gi}{\def\PYG@tc##1{\textcolor[rgb]{0.00,0.52,0.00}{##1}}}
\@namedef{PYG@tok@gr}{\def\PYG@tc##1{\textcolor[rgb]{0.89,0.00,0.00}{##1}}}
\@namedef{PYG@tok@ge}{\let\PYG@it=\textit}
\@namedef{PYG@tok@gs}{\let\PYG@bf=\textbf}
\@namedef{PYG@tok@gp}{\let\PYG@bf=\textbf\def\PYG@tc##1{\textcolor[rgb]{0.00,0.00,0.50}{##1}}}
\@namedef{PYG@tok@go}{\def\PYG@tc##1{\textcolor[rgb]{0.44,0.44,0.44}{##1}}}
\@namedef{PYG@tok@gt}{\def\PYG@tc##1{\textcolor[rgb]{0.00,0.27,0.87}{##1}}}
\@namedef{PYG@tok@err}{\def\PYG@bc##1{{\setlength{\fboxsep}{\string -\fboxrule}\fcolorbox[rgb]{1.00,0.00,0.00}{1,1,1}{\strut ##1}}}}
\@namedef{PYG@tok@kc}{\let\PYG@bf=\textbf\def\PYG@tc##1{\textcolor[rgb]{0.00,0.50,0.00}{##1}}}
\@namedef{PYG@tok@kd}{\let\PYG@bf=\textbf\def\PYG@tc##1{\textcolor[rgb]{0.00,0.50,0.00}{##1}}}
\@namedef{PYG@tok@kn}{\let\PYG@bf=\textbf\def\PYG@tc##1{\textcolor[rgb]{0.00,0.50,0.00}{##1}}}
\@namedef{PYG@tok@kr}{\let\PYG@bf=\textbf\def\PYG@tc##1{\textcolor[rgb]{0.00,0.50,0.00}{##1}}}
\@namedef{PYG@tok@bp}{\def\PYG@tc##1{\textcolor[rgb]{0.00,0.50,0.00}{##1}}}
\@namedef{PYG@tok@fm}{\def\PYG@tc##1{\textcolor[rgb]{0.00,0.00,1.00}{##1}}}
\@namedef{PYG@tok@vc}{\def\PYG@tc##1{\textcolor[rgb]{0.10,0.09,0.49}{##1}}}
\@namedef{PYG@tok@vg}{\def\PYG@tc##1{\textcolor[rgb]{0.10,0.09,0.49}{##1}}}
\@namedef{PYG@tok@vi}{\def\PYG@tc##1{\textcolor[rgb]{0.10,0.09,0.49}{##1}}}
\@namedef{PYG@tok@vm}{\def\PYG@tc##1{\textcolor[rgb]{0.10,0.09,0.49}{##1}}}
\@namedef{PYG@tok@sa}{\def\PYG@tc##1{\textcolor[rgb]{0.73,0.13,0.13}{##1}}}
\@namedef{PYG@tok@sb}{\def\PYG@tc##1{\textcolor[rgb]{0.73,0.13,0.13}{##1}}}
\@namedef{PYG@tok@sc}{\def\PYG@tc##1{\textcolor[rgb]{0.73,0.13,0.13}{##1}}}
\@namedef{PYG@tok@dl}{\def\PYG@tc##1{\textcolor[rgb]{0.73,0.13,0.13}{##1}}}
\@namedef{PYG@tok@s2}{\def\PYG@tc##1{\textcolor[rgb]{0.73,0.13,0.13}{##1}}}
\@namedef{PYG@tok@sh}{\def\PYG@tc##1{\textcolor[rgb]{0.73,0.13,0.13}{##1}}}
\@namedef{PYG@tok@s1}{\def\PYG@tc##1{\textcolor[rgb]{0.73,0.13,0.13}{##1}}}
\@namedef{PYG@tok@mb}{\def\PYG@tc##1{\textcolor[rgb]{0.40,0.40,0.40}{##1}}}
\@namedef{PYG@tok@mf}{\def\PYG@tc##1{\textcolor[rgb]{0.40,0.40,0.40}{##1}}}
\@namedef{PYG@tok@mh}{\def\PYG@tc##1{\textcolor[rgb]{0.40,0.40,0.40}{##1}}}
\@namedef{PYG@tok@mi}{\def\PYG@tc##1{\textcolor[rgb]{0.40,0.40,0.40}{##1}}}
\@namedef{PYG@tok@il}{\def\PYG@tc##1{\textcolor[rgb]{0.40,0.40,0.40}{##1}}}
\@namedef{PYG@tok@mo}{\def\PYG@tc##1{\textcolor[rgb]{0.40,0.40,0.40}{##1}}}
\@namedef{PYG@tok@ch}{\let\PYG@it=\textit\def\PYG@tc##1{\textcolor[rgb]{0.24,0.48,0.48}{##1}}}
\@namedef{PYG@tok@cm}{\let\PYG@it=\textit\def\PYG@tc##1{\textcolor[rgb]{0.24,0.48,0.48}{##1}}}
\@namedef{PYG@tok@cpf}{\let\PYG@it=\textit\def\PYG@tc##1{\textcolor[rgb]{0.24,0.48,0.48}{##1}}}
\@namedef{PYG@tok@c1}{\let\PYG@it=\textit\def\PYG@tc##1{\textcolor[rgb]{0.24,0.48,0.48}{##1}}}
\@namedef{PYG@tok@cs}{\let\PYG@it=\textit\def\PYG@tc##1{\textcolor[rgb]{0.24,0.48,0.48}{##1}}}

\def\PYGZus{\char`\_}
\def\PYGZob{\char`\{}
\def\PYGZcb{\char`\}}

\def\PYGZlt{\char`\<}
\def\PYGZgt{\char`\>}

\def\PYGZhy{\char`\-}

\def\PYGZdq{\char`\"}

\makeatother

\makeatletter
\def\PYGdefault@reset{\let\PYGdefault@it=\relax \let\PYGdefault@bf=\relax%
    \let\PYGdefault@ul=\relax \let\PYGdefault@tc=\relax%
    \let\PYGdefault@bc=\relax \let\PYGdefault@ff=\relax}
\def\PYGdefault@tok#1{\csname PYGdefault@tok@#1\endcsname}
\def\PYGdefault@toks#1+{\ifx\relax#1\empty\else%
    \PYGdefault@tok{#1}\expandafter\PYGdefault@toks\fi}
\def\PYGdefault@do#1{\PYGdefault@bc{\PYGdefault@tc{\PYGdefault@ul{%
    \PYGdefault@it{\PYGdefault@bf{\PYGdefault@ff{#1}}}}}}}
\def\PYGdefault#1#2{\PYGdefault@reset\PYGdefault@toks#1+\relax+\PYGdefault@do{#2}}

\@namedef{PYGdefault@tok@w}{\def\PYGdefault@tc##1{\textcolor[rgb]{0.73,0.73,0.73}{##1}}}
\@namedef{PYGdefault@tok@c}{\let\PYGdefault@it=\textit\def\PYGdefault@tc##1{\textcolor[rgb]{0.24,0.48,0.48}{##1}}}
\@namedef{PYGdefault@tok@cp}{\def\PYGdefault@tc##1{\textcolor[rgb]{0.61,0.40,0.00}{##1}}}
\@namedef{PYGdefault@tok@k}{\let\PYGdefault@bf=\textbf\def\PYGdefault@tc##1{\textcolor[rgb]{0.00,0.50,0.00}{##1}}}
\@namedef{PYGdefault@tok@kp}{\def\PYGdefault@tc##1{\textcolor[rgb]{0.00,0.50,0.00}{##1}}}
\@namedef{PYGdefault@tok@kt}{\def\PYGdefault@tc##1{\textcolor[rgb]{0.69,0.00,0.25}{##1}}}
\@namedef{PYGdefault@tok@o}{\def\PYGdefault@tc##1{\textcolor[rgb]{0.40,0.40,0.40}{##1}}}
\@namedef{PYGdefault@tok@ow}{\let\PYGdefault@bf=\textbf\def\PYGdefault@tc##1{\textcolor[rgb]{0.67,0.13,1.00}{##1}}}
\@namedef{PYGdefault@tok@nb}{\def\PYGdefault@tc##1{\textcolor[rgb]{0.00,0.50,0.00}{##1}}}
\@namedef{PYGdefault@tok@nf}{\def\PYGdefault@tc##1{\textcolor[rgb]{0.00,0.00,1.00}{##1}}}
\@namedef{PYGdefault@tok@nc}{\let\PYGdefault@bf=\textbf\def\PYGdefault@tc##1{\textcolor[rgb]{0.00,0.00,1.00}{##1}}}
\@namedef{PYGdefault@tok@nn}{\let\PYGdefault@bf=\textbf\def\PYGdefault@tc##1{\textcolor[rgb]{0.00,0.00,1.00}{##1}}}
\@namedef{PYGdefault@tok@ne}{\let\PYGdefault@bf=\textbf\def\PYGdefault@tc##1{\textcolor[rgb]{0.80,0.25,0.22}{##1}}}
\@namedef{PYGdefault@tok@nv}{\def\PYGdefault@tc##1{\textcolor[rgb]{0.10,0.09,0.49}{##1}}}
\@namedef{PYGdefault@tok@no}{\def\PYGdefault@tc##1{\textcolor[rgb]{0.53,0.00,0.00}{##1}}}
\@namedef{PYGdefault@tok@nl}{\def\PYGdefault@tc##1{\textcolor[rgb]{0.46,0.46,0.00}{##1}}}
\@namedef{PYGdefault@tok@ni}{\let\PYGdefault@bf=\textbf\def\PYGdefault@tc##1{\textcolor[rgb]{0.44,0.44,0.44}{##1}}}
\@namedef{PYGdefault@tok@na}{\def\PYGdefault@tc##1{\textcolor[rgb]{0.41,0.47,0.13}{##1}}}
\@namedef{PYGdefault@tok@nt}{\let\PYGdefault@bf=\textbf\def\PYGdefault@tc##1{\textcolor[rgb]{0.00,0.50,0.00}{##1}}}
\@namedef{PYGdefault@tok@nd}{\def\PYGdefault@tc##1{\textcolor[rgb]{0.67,0.13,1.00}{##1}}}
\@namedef{PYGdefault@tok@s}{\def\PYGdefault@tc##1{\textcolor[rgb]{0.73,0.13,0.13}{##1}}}
\@namedef{PYGdefault@tok@sd}{\let\PYGdefault@it=\textit\def\PYGdefault@tc##1{\textcolor[rgb]{0.73,0.13,0.13}{##1}}}
\@namedef{PYGdefault@tok@si}{\let\PYGdefault@bf=\textbf\def\PYGdefault@tc##1{\textcolor[rgb]{0.64,0.35,0.47}{##1}}}
\@namedef{PYGdefault@tok@se}{\let\PYGdefault@bf=\textbf\def\PYGdefault@tc##1{\textcolor[rgb]{0.67,0.36,0.12}{##1}}}
\@namedef{PYGdefault@tok@sr}{\def\PYGdefault@tc##1{\textcolor[rgb]{0.64,0.35,0.47}{##1}}}
\@namedef{PYGdefault@tok@ss}{\def\PYGdefault@tc##1{\textcolor[rgb]{0.10,0.09,0.49}{##1}}}
\@namedef{PYGdefault@tok@sx}{\def\PYGdefault@tc##1{\textcolor[rgb]{0.00,0.50,0.00}{##1}}}
\@namedef{PYGdefault@tok@m}{\def\PYGdefault@tc##1{\textcolor[rgb]{0.40,0.40,0.40}{##1}}}
\@namedef{PYGdefault@tok@gh}{\let\PYGdefault@bf=\textbf\def\PYGdefault@tc##1{\textcolor[rgb]{0.00,0.00,0.50}{##1}}}
\@namedef{PYGdefault@tok@gu}{\let\PYGdefault@bf=\textbf\def\PYGdefault@tc##1{\textcolor[rgb]{0.50,0.00,0.50}{##1}}}
\@namedef{PYGdefault@tok@gd}{\def\PYGdefault@tc##1{\textcolor[rgb]{0.63,0.00,0.00}{##1}}}
\@namedef{PYGdefault@tok@gi}{\def\PYGdefault@tc##1{\textcolor[rgb]{0.00,0.52,0.00}{##1}}}
\@namedef{PYGdefault@tok@gr}{\def\PYGdefault@tc##1{\textcolor[rgb]{0.89,0.00,0.00}{##1}}}
\@namedef{PYGdefault@tok@ge}{\let\PYGdefault@it=\textit}
\@namedef{PYGdefault@tok@gs}{\let\PYGdefault@bf=\textbf}
\@namedef{PYGdefault@tok@gp}{\let\PYGdefault@bf=\textbf\def\PYGdefault@tc##1{\textcolor[rgb]{0.00,0.00,0.50}{##1}}}
\@namedef{PYGdefault@tok@go}{\def\PYGdefault@tc##1{\textcolor[rgb]{0.44,0.44,0.44}{##1}}}
\@namedef{PYGdefault@tok@gt}{\def\PYGdefault@tc##1{\textcolor[rgb]{0.00,0.27,0.87}{##1}}}
\@namedef{PYGdefault@tok@err}{\def\PYGdefault@bc##1{{\setlength{\fboxsep}{\string -\fboxrule}\fcolorbox[rgb]{1.00,0.00,0.00}{1,1,1}{\strut ##1}}}}
\@namedef{PYGdefault@tok@kc}{\let\PYGdefault@bf=\textbf\def\PYGdefault@tc##1{\textcolor[rgb]{0.00,0.50,0.00}{##1}}}
\@namedef{PYGdefault@tok@kd}{\let\PYGdefault@bf=\textbf\def\PYGdefault@tc##1{\textcolor[rgb]{0.00,0.50,0.00}{##1}}}
\@namedef{PYGdefault@tok@kn}{\let\PYGdefault@bf=\textbf\def\PYGdefault@tc##1{\textcolor[rgb]{0.00,0.50,0.00}{##1}}}
\@namedef{PYGdefault@tok@kr}{\let\PYGdefault@bf=\textbf\def\PYGdefault@tc##1{\textcolor[rgb]{0.00,0.50,0.00}{##1}}}
\@namedef{PYGdefault@tok@bp}{\def\PYGdefault@tc##1{\textcolor[rgb]{0.00,0.50,0.00}{##1}}}
\@namedef{PYGdefault@tok@fm}{\def\PYGdefault@tc##1{\textcolor[rgb]{0.00,0.00,1.00}{##1}}}
\@namedef{PYGdefault@tok@vc}{\def\PYGdefault@tc##1{\textcolor[rgb]{0.10,0.09,0.49}{##1}}}
\@namedef{PYGdefault@tok@vg}{\def\PYGdefault@tc##1{\textcolor[rgb]{0.10,0.09,0.49}{##1}}}
\@namedef{PYGdefault@tok@vi}{\def\PYGdefault@tc##1{\textcolor[rgb]{0.10,0.09,0.49}{##1}}}
\@namedef{PYGdefault@tok@vm}{\def\PYGdefault@tc##1{\textcolor[rgb]{0.10,0.09,0.49}{##1}}}
\@namedef{PYGdefault@tok@sa}{\def\PYGdefault@tc##1{\textcolor[rgb]{0.73,0.13,0.13}{##1}}}
\@namedef{PYGdefault@tok@sb}{\def\PYGdefault@tc##1{\textcolor[rgb]{0.73,0.13,0.13}{##1}}}
\@namedef{PYGdefault@tok@sc}{\def\PYGdefault@tc##1{\textcolor[rgb]{0.73,0.13,0.13}{##1}}}
\@namedef{PYGdefault@tok@dl}{\def\PYGdefault@tc##1{\textcolor[rgb]{0.73,0.13,0.13}{##1}}}
\@namedef{PYGdefault@tok@s2}{\def\PYGdefault@tc##1{\textcolor[rgb]{0.73,0.13,0.13}{##1}}}
\@namedef{PYGdefault@tok@sh}{\def\PYGdefault@tc##1{\textcolor[rgb]{0.73,0.13,0.13}{##1}}}
\@namedef{PYGdefault@tok@s1}{\def\PYGdefault@tc##1{\textcolor[rgb]{0.73,0.13,0.13}{##1}}}
\@namedef{PYGdefault@tok@mb}{\def\PYGdefault@tc##1{\textcolor[rgb]{0.40,0.40,0.40}{##1}}}
\@namedef{PYGdefault@tok@mf}{\def\PYGdefault@tc##1{\textcolor[rgb]{0.40,0.40,0.40}{##1}}}
\@namedef{PYGdefault@tok@mh}{\def\PYGdefault@tc##1{\textcolor[rgb]{0.40,0.40,0.40}{##1}}}
\@namedef{PYGdefault@tok@mi}{\def\PYGdefault@tc##1{\textcolor[rgb]{0.40,0.40,0.40}{##1}}}
\@namedef{PYGdefault@tok@il}{\def\PYGdefault@tc##1{\textcolor[rgb]{0.40,0.40,0.40}{##1}}}
\@namedef{PYGdefault@tok@mo}{\def\PYGdefault@tc##1{\textcolor[rgb]{0.40,0.40,0.40}{##1}}}
\@namedef{PYGdefault@tok@ch}{\let\PYGdefault@it=\textit\def\PYGdefault@tc##1{\textcolor[rgb]{0.24,0.48,0.48}{##1}}}
\@namedef{PYGdefault@tok@cm}{\let\PYGdefault@it=\textit\def\PYGdefault@tc##1{\textcolor[rgb]{0.24,0.48,0.48}{##1}}}
\@namedef{PYGdefault@tok@cpf}{\let\PYGdefault@it=\textit\def\PYGdefault@tc##1{\textcolor[rgb]{0.24,0.48,0.48}{##1}}}
\@namedef{PYGdefault@tok@c1}{\let\PYGdefault@it=\textit\def\PYGdefault@tc##1{\textcolor[rgb]{0.24,0.48,0.48}{##1}}}
\@namedef{PYGdefault@tok@cs}{\let\PYGdefault@it=\textit\def\PYGdefault@tc##1{\textcolor[rgb]{0.24,0.48,0.48}{##1}}}

\makeatother

\title{Efficient comparison of sentence embeddings}

\author{
  Spyros Zoupanos \\
  Department of Informatics,\\ 
  Ionian University, \\
  Corfu, Greece\\
  \texttt{spyros@zoupanos.net} \\
  \And
  Stratis Kolovos \\
  Department of Information and\\Communication Systems Engineering, \\
  University of the Aegean, \\
  Samos, Greece\\
  \texttt{icsdd20018@icsd.aegean.gr} \\
  \And
  Athanasios Kanavos \\
  Department of Information and\\Communication Systems Engineering, \\
  University of the Aegean, \\
  Samos, Greece\\
  \texttt{icsdd20017@icsd.aegean.gr} \\
  \And
  Orestis Papadimitriou \\
  Department of Information and\\Communication Systems Engineering, \\
  University of the Aegean, \\
  Samos, Greece\\
  \texttt{icsdd20016@icsd.aegean.gr} \\  
  \And
  Manolis Maragoudakis \\
  Department of Informatics,\\ 
  Ionian University, \\
  Corfu, Greece\\
  \texttt{mmarag@ionio.gr} \\
}

\begin{document}
\maketitle

\begin{abstract}
The domain of natural language processing (NLP), which has greatly evolved over the last years, has highly benefited from the recent developments in word and sentence embeddings. Such embeddings enable the transformation of complex NLP tasks, like semantic similarity or Question and Answering (Q\&A), into much simpler to perform vector comparisons. However, such a problem transformation raises new challenges like the efficient comparison of embeddings and their manipulation. In this work, we will discuss about various word and sentence embeddings algorithms, we will select a sentence embedding algorithm, BERT, as our algorithm of choice and we will evaluate the performance of two vector comparison approaches, FAISS and Elasticsearch, in the specific problem of sentence embeddings. According to the results, FAISS outperforms Elasticsearch when used in a centralized environment with only one node, especially when big datasets are included.
\end{abstract}


\section{Introduction}
Natural language processing (NLP) focuses on the interactions between computers and human language and, in particular, on how to program computers to efficiently process language data. NLP has greatly evolved the latest years making the processing of millions of web pages by Google and its rivals to seem like a trivial task which can be completed in less than a second, while a few years ago, the analysis of one sentence could last several minutes~\cite{cambria2014icim}. This great improvement in NLP performance makes it possible to use computers for various complex tasks like machine translation, dialogue systems, sentiment analysis, part-of-speech tagging and many more.

In this work we will investigate the problem of natural language processing using word and sentence embeddings in the big data era. We present various embedding algorithms and we select one of the most prominent algorithms, BERT, to investigate the problem of efficient comparison of sentence embeddings. We will focus on two infrastructure choices for such a comparison. The available approaches will be analyzed and their performance on sentence embedding selection will be presented and compared against a third, baseline approach. The approaches will also be compared among them and conclusions will be drawn given the available datasets and resources.

\section{Word and sentence embeddings}
One of the biggest challenges in NLP is the problem of high-di-mensionality and data sparseness which can be addressed with the use of distributed vectors of word embeddings. The key idea of this approach is that words with similar meaning tend to appear in a similar context. Therefore, the vectors are constructed in a way that they encapsulate the characteristics of the neighbors of a word. Comparison among vectors is performed using metrics such as cosine similarity, demonstrating the similarity of the words that the vectors correspond to.

These embeddings can be used as the first data processing step in a deep learning model. They are usually pre-trained in an unlabeled corpus~\cite{mikolov2013arxiv, mikolov2013nips} where they can capture syntactical and semantic information. Because of their small dimensionality and due to the fact that these are dense vectors, they can be ideal for processing core NLP tasks.

Shallow neural networks have been the models to create such embeddings over the years with good performance. Deep learning NLP models use them also to represent their words, phrases and sentences, making these embeddings one of the key differences between traditional word-based models and deep learning-based models. Word embeddings have been used for a wide range of NLP tasks with significant results~\cite{weston2011ijcai, socher2011icml, turney2010jair, cambria2017iis}.

\subsection{Word-level encodings}
Mikolov et al.~\cite{mikolov2013arxiv, mikolov2013nips} transformed word embedding algorithms by proposing the CBOW and skip-gram models. Both of these models work on a window of words, but they follow different approaches. CBOW calculates the conditional probability of a target word given the context words around it within a window of specific size. On the contrary, the skip-gram model predicts the context words within a window of words given a specific target word. Word2vec uses a large corpus of text as its input, creating a vector space that typically has several hundred dimensions, with each unique word being assigned a vector in the space. As a result, word vectors in the vector space are positioned so that words sharing the same context in the corpus are close to one another.

FastText~\cite{joulin2019arxiv} is exploring ways to make linear text classifiers like\cite{joachims1998ecml, mccallum1998aaai, fan2008jmlr} to train to billion of words within a few minutes and to achieve performance comparable to the state-of-the-art models. It is inspired by Word2Vec~\cite{mikolov2013arxiv} with its main difference being that it focuses on n-grams instead of words to make more fine grained representations.


GloVe~\cite{pennington2014emnlp} is another algorithm creating word vectors. The authors have developed a global log-bilinear regression model combining the benefits of two major families of models in the literature: global matrix factorization and local context window models.
They claim that matrix factorization methods manage to outperform shallow window-based methods like Word2Vec in scenarios where there is a vast amount of data repetition.


\subsection{Contextualized Word Embeddings}
Traditional word embedding methods did not take into account the fact that a given word can have completely different meaning in different contexts. Being able to understand and differentiate among different senses of a word increases the quality of the produced results. Therefore, taking into account syntactical information and handling polysemic behavior leads to better word representations. Recent approaches take the above into account and calculate word representations as a function of the word's contexts.

Word2Vec algorithm, which was mentioned in the previous section, as well as other traditional word embedding methods, create a global vector representation of a word, considering all the sentences where the word is present. New models, like Embedding from Language Model (ELMo)~\cite{peters2018arxiv}, propose contextual word embeddings. More specifically, for every context where a word is used, ELMo produces a word embedding allowing to have different representations for different senses of the same word.

\subsection{Sentence Embeddings}
BERT algorithm that has been recently proposed by Devlin et al.~\cite{devlin2019nacacl} introduces the idea of relationship among sentences. It uses a transformer network to pre-train a language model to extract contextual word embeddings. Its language modeling is based on different tasks. In one task, BERT randomly masks a percentage of words in the sentences and only predicts the masked words while in another task, it tries to predict the next sentence given a specific sentence. The latter task tries to model the relationship among sentences. BERT outperforms state-of-the-art techniques by a large margin on key NLP tasks such as Question answering (QA) and Natural Language Inference (NLI) where understanding relations among sentences is very important. 

Roberta~\cite{liu2019arxiv} focuses on improving the training of the BERT model. Its authors claim that BERT is under-trained and propose four modifications to improve its performance and outperform many models published since it. They propose to train the model longer with bigger batches and longer sequences, to remove the next sentence objective and to dynamically change the masking patters applied to the training data.

%

A sentence-BERT~\cite{reimers2019arxiv} algorithm may well be a modification of the BERT algorithm that uses siamese and triplet networks to derive semantically meaningful sentence embeddings. Additionally to semantic similarity comparisons, clustering and knowledge retrieval via semantic search can be very efficiently performed on modern hardware, making SBERT suitable for innovative tasks not previously possible for BERT.

Two of the limitations of BERT is that it assumes that the predicted tokens are independent of each other given the unmasked tokens and that it suffers from pretrain-finetune discrepancy. XLNet~\cite{yang2019arxiv} is proposed to overcome these limitations. It enables learning bidirectional context which maximizes expected likelihood over all permutations of the factorization and it overcomes BERT with regard to auto-regressive formulation. It also integrates into its pre-training ideas from Transformer-XL, a state-of-the-art autoregressive model. The result is that XLNet outperforms BERT in various tasks like question answering, natural language inference, sentiment analysis, and document ranking.


ALBERT~\cite{lan2019corr} proposes two optimizations to address memory limitations and training times that occur because of necessary model increases of BERT. Both of the optimizations focus on the reduction of BERT parameters without affecting performance: (1) they decompose the large vocabulary embedding matrix into smaller matrices and (2) they share parameters across different layers. Their result is an increased training performance, a great reduction in number of used parameters (18x) and a more stable training phase.

\section{Embedding comparison frameworks}
\begin{figure*}
  \includegraphics[width=\linewidth]{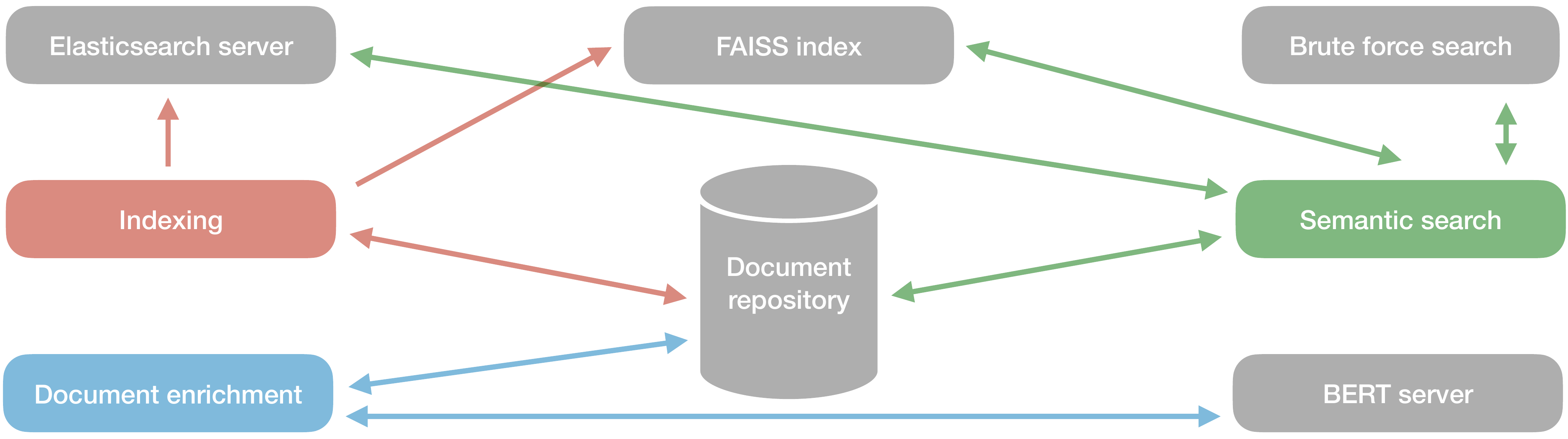}
  \caption{An architecture overview. We have highlighted three different modules corresponding to the three different phases of our experimental setup: (a) a \textit{document enrichment} module which enriches the documents and their corresponding queries with embeddings retrieved from the BERT server, (b) an \textit{indexing} module that contacts the Elasticsearch server and the FAISS index for document and embedding indexing and (c) a \textit{semantic search} module which uses the query embeddings, and the document embeddings in the case of the naive/brute-force approach, to retrieve the needed results.}
  \label{fig:FAISS}
\end{figure*}

The construction of high-quality word and sentence embeddings is crucial to achieve good results in various NLP tasks. However, an efficient and scalable way of comparing them is also very important. In this section, we will investigate two different frameworks (two different approaches) to compare large number of such embeddings.

\subsection{Elasticsearch}


Elasticsearch~\cite{elasticsearch} is an open source, highly scalable, full-text search engine  which is written in Java. An Elasticsearch cluster can be comprised of one or more Elasticsearch nodes, which are Elasticsearch server instances.

The Elasticsearch engine was initially written in Java to support free and open-source information retrieval from software libraries. As a Java library it was cumbersome to be used and therefore an abstraction layer called Compass was developed. In later years, the libraries were rewritten to ensure real-time performance and a standalone Elasticsearch server was released.

Elasticsearch uses the REST API to carry out its operations and is multitenant~\cite{kononenko2014msr}. The lightweight JSON (JavaScript object notation)-queries are accessed using the basic HTTP interface. Elasticsearch shares many analogies with RDBMS. It stores documents in JSON format which is the equivalent of a row of a RDBMS table. Each document consists of fields which can be primitive types or complex structures. Even if documents are considered schema-free, they have a document type associated with them. Document type is the equivalent of a table of a RDBMS since it describes the fields that can be specified for a particular document. However, documents of the same type may have different schemas. All the documents in Elasticsearch are stored in indices permitting to store, update and query them. The index is the equivalent of a RDBMS database.

Elasticsearch index consists of one or more Apache Lucene~\cite{lucene} indexes which are called shards. While the number of shards of an index is predefined before index creation, documents mapping to shards, responsible for their storage and indexing, depends on load. This load balancing permits all shards to be used simultaneously improving performance. Elasticsearch improves even more its performance and scalability by the automatic re-distribution of shards based on the available nodes. Irrespectively of the number of shards of an index and its internal load balancing, it is visible to the client as a single structure.

One of the advantages of Elasticsearch is its horizontal (sharding) and vertical (powerful servers) scalability. Horizontal scalability is ensured by distributing automatically index shards across nodes and ensures that are loaded equally.

Elasticsearch outstands also for its flexibility. Since it doesn't impose any schema to the documents stored, it can easily accommodate data that changes over time. When a new document is added with more fields than the already loaded documents, Elasticsearch will update the index mapping without affecting what has already been loaded. Moreover, Elasticsearch's index design, which is based on a collection of shards, ensures correct load balancing of index update tasks that happen when new data arrives. RDBMS have a predefined schema that can not be easily updated after data loading. Moreover, index updates in the presence of large amount of data is not a trivial task. 

Last but not least, since Elasticsearch operates on documents, it doesn't need to perform any normalization process of the data. Normalization process facilitates the create, update and remove operations. However, it is very likely to result to more costly read operations since statements may need data from different tables that should be combined (joined) to provide the final result.

\subsection{FAISS}
FAISS~\cite{johnson2021tbg} addresses the problem of efficient comparison and similarity search of high-dimensional vectors (500 to 1,000+ dimensions). Such vector representations are frequently used in classification of images, videos or text. In the context of text comparison, word and sentence embeddings are represented as large vectors of numbers. One of the most expensive operations in this setting is to compute a k-NN (k-Nearest Neighbor) graph, which is a directed graph where each vector of the database is a node and each edge connects a node to its k nearest neighbors.

The authors of~\cite{johnson2021tbg} focus on how GPUs can be efficiently utilized in the problem of high-dimensional similarity comparison. They have developed and present a GPU k-selection algorithm which performs approximate and exact k-neighbor search. They make a complexity analysis and present a range of experiments showing that the proposed improvements outperform the state of the art algorithms in exact and approximate search comparisons.

More specifically, in the exact search setting, they compare against other algorithms in k-selection performance where FAISS outperforms competitors for various values of k. Another exact-search comparison performed is k-means clustering where, in a large scale setup, they perform comparable performance to the competition but without any pre-processing. The authors also present good results in exact nearest search. Another domain where FAISS excels is the approximate nearest neighbor search on large datasets where they demonstrate better accuracy and lower execution times against other implementations~\cite{wieschollek2016cvpr}.

\section{Embedding comparison setup}

\begin{listing}
\begin{center}
\begin{minipage}{.60\textwidth}
\begin{Verbatim}[commandchars=\\\{\}, frame=single, framesep=1mm, xleftmargin=11pt, tabsize=1]
\PYG{p}{\PYGZob{}}
  \PYG{l+s+s2}{\PYGZdq{}document\PYGZus{}embeddings\PYGZdq{}}\PYG{o}{:} \PYG{p}{[}\PYG{l+m+mf}{0.178529500961}\PYG{p}{,} \PYG{p}{...],}
  \PYG{l+s+s2}{\PYGZdq{}document\PYGZus{}text\PYGZdq{}}\PYG{o}{:} \PYG{l+s+s2}{\PYGZdq{}Email marketing \PYGZhy{} Wikipedia}
\PYG{l+s+s2}{                      \PYGZlt{}H1\PYGZgt{}Email marketing\PYGZlt{}/H1\PYGZgt{}}
\PYG{l+s+s2}{                      Jump to...\PYGZdq{}}\PYG{p}{,}
  \PYG{l+s+s2}{\PYGZdq{}example\PYGZus{}id\PYGZdq{}}\PYG{o}{:} \PYG{l+m+mf}{5655493461695504401}\PYG{p}{,}
  \PYG{l+s+s2}{\PYGZdq{}question\PYGZus{}embeddings\PYGZdq{}}\PYG{o}{:} \PYG{p}{[}\PYG{o}{\PYGZhy{}}\PYG{l+m+mf}{0.051026359897}\PYG{p}{,}
                          \PYG{p}{...],}
  \PYG{l+s+s2}{\PYGZdq{}question\PYGZus{}text\PYGZdq{}}\PYG{o}{:}  \PYG{l+s+s2}{\PYGZdq{}which is the most common}
\PYG{l+s+s2}{                      use of opt\PYGZhy{}in e\PYGZhy{}mail}
\PYG{l+s+s2}{                      marketing\PYGZdq{}}
\PYG{p}{\PYGZcb{}}
\end{Verbatim}
\end{minipage}
\end{center}
\caption{An example of an enriched JSON dictionary with BERT embeddings of the Google NQ dataset.}
\label{listing:enriched_line}
\end{listing}

In this section we will perform benchmarks using Google's Natural Questions (NQ) dataset~\cite{kwiatkowski2019tacl}. We will first give an overview of the dataset and then we will present two different benchmarks: (a) We will compare the performance of three different approaches for vector comparison: naive/brute-force, FAISS and Elasticsearch. (b) Then we will perform an evaluation of the quality of the embeddings based on the questions asked and the documents that correspond to these questions.

\subsection{Dataset overview}
In Artificial Intelligence and more specifically, in Information Retrieval, the goal is to create systems that can read the web and answer complex questions on any topic. Question-answering systems can have a significant impact on how we interact with information. The ability to understand and answer questions about a given content is also something we generally associate with intelligence, which is why open-domain question answering is one of the benchmark tasks for Artificial Intelligence development.
As part of the natural questions corpus, query answering systems are required to read and comprehend an entire Wikipedia article that may or may not contain the answer to the question. 
What differentiates Google's Natural Questions (NQ)~\cite{kwiatkowski2019tacl} dataset from previous Query Answering (QA) datasets is the fact that it includes real user questions that may need the full document to be read to return a valid answer. This fact makes it a more realistic and challenging dataset to evaluate compared to previous efforts. It contains real questions and answers gathered from Google search, as well as annotations from Wikipedia. NQ was designed for testing and training automated question answering systems.

In the sequel, we will focus on the simplified version of the dataset that is composed of 307,373 examples that contain a simplified version of a Wikipedia web-page, a question/query that should be answered from the web-page, long answer candidates and the answer of one annotator on the query asked based on that specific page. The annotator provides a long and short answer based on the document, wherever possible.
In~\cite{kwiatkowski2019tacl} the authors have established high human upper bounds for long and short answers (90\% precision, 85\% recall and 79\% precision, 72\% recall respectively).

\subsection{Dataset enrichment using sentence embeddings}
~\label{sec:dataset_enrichment}
We use the Google NQ dataset presented above for our benchmarks which we enrich with vector embeddings. More specifically, the simplified version of the dataset is composed of 307,373 lines where each line is a separate JSON string. The latter is a dictionary with the following keys: \textit{annotations} (containing the proposed long and short answers by the annotator), \textit{document\_text}, \textit{document\_url}, \textit{example\_id}, \textit{long\_answer\_candidates} and \textit{question\_text}. From these keys, we retain the document, the corresponding query and, for reference reasons, the id of the example. We create embeddings for the document and the query using BERT and the model \textit{uncased\_L-12\_H-768\_A-12}. We do not pre-train the model before generating the needed embeddings. The resulting embeddings are \textit{vectors containing 768 numbers} and a dictionary of the procedure described above is depicted at Listing~\ref{listing:enriched_line}.

The Google NQ dataset contains hundreds of thousands of lines and the processing described above needs special care since it is a time and resource consuming procedure. Therefore, the dataset is read in batches of lines, and each batch is processed independently. The documents and queries of each batch are transferred to BERT which runs as a service in a docker container. For efficiency reasons, we have defined the maximum length of a sequence to 256. The number of workers is set to 1. As soon as the document and query embeddings of a batch are generated, then lines as presented at Listing~\ref{listing:enriched_line} are constructed and stored in a separate file resulting to the enriched dataset which will be used in the following benchmarks.

Following the aforementioned procedure, we generated 100,000 lines (similar to the one presented at Listing~\ref{listing:enriched_line}) in 24 hours. For the dataset generation we used a virtual machine with 8GB of RAM and 6-cores of an AMD Ryzen 7 PRO 5850U CPU. The BERT serving instantiation used was at version 1.9.8.

\begin{figure*}
    \centering
    \subfloat[\centering Log scale comparison of the Naive, FAISS and Elasticsearch approaches.]{{\includegraphics[width=.465\textwidth]{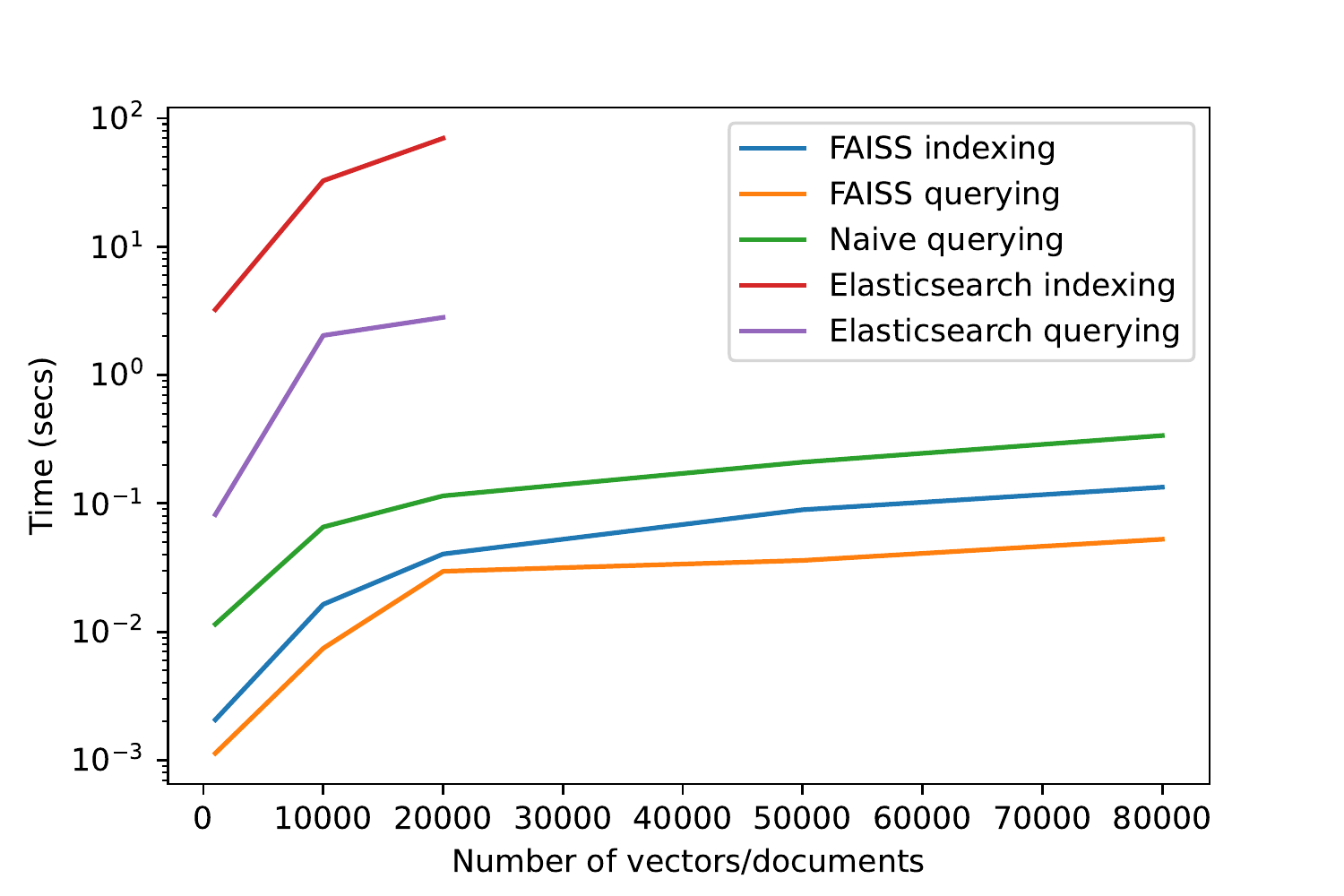} }}%
    \qquad
    \subfloat[\centering Normal scale comparison of FAISS and Naive approaches.]{{\includegraphics[width=.465\textwidth]{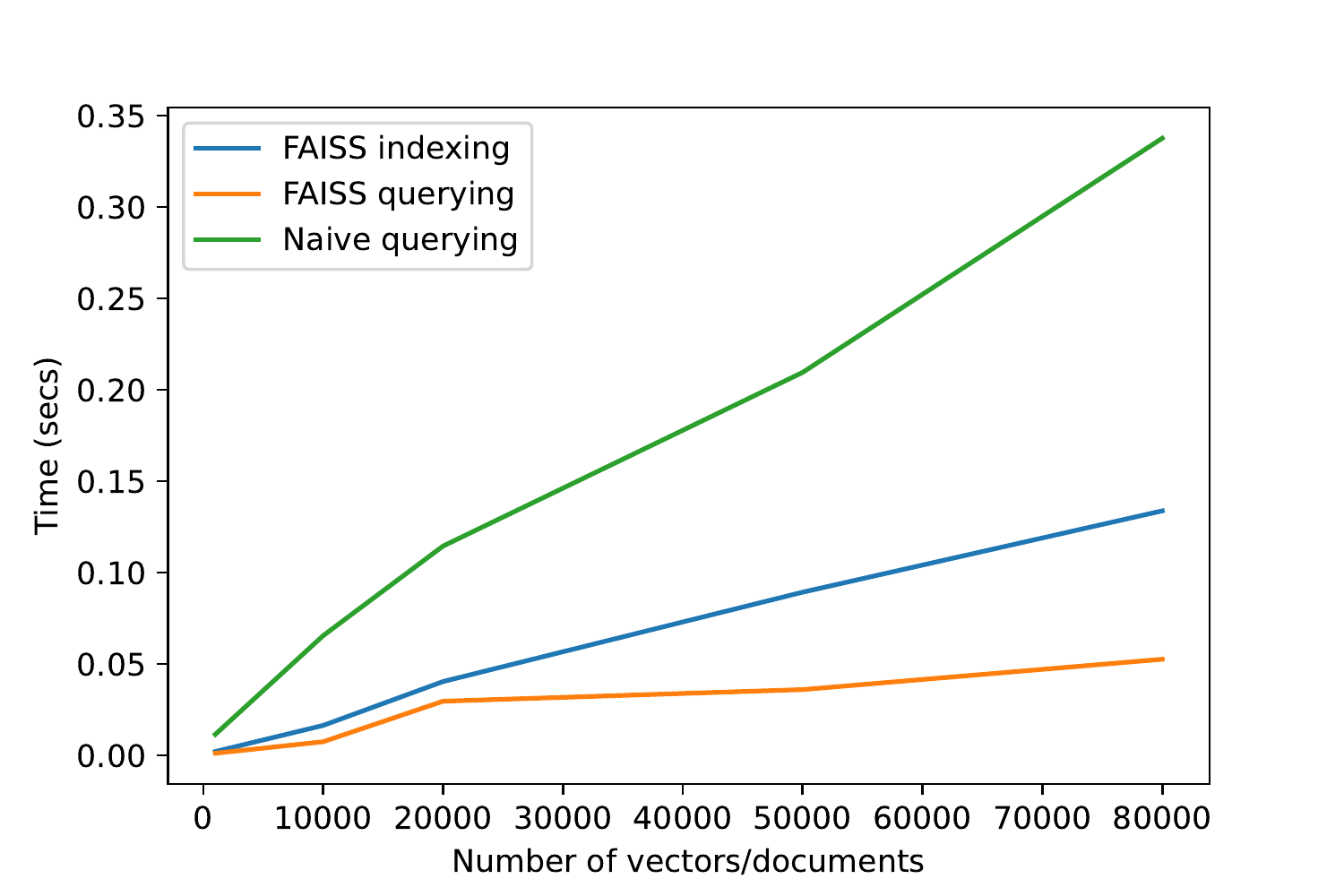} }}%
    \caption{Performance of three vector comparison approaches. For FAISS and Elasticsearch we show the indexing and querying performance for various dataset sizes while for the naive/brute-force search, we only show the querying performance.}%
    \label{fig:naive_faiss_elastic_comparison}%
\end{figure*}

\subsection{Efficient vector comparison}
We use three different approaches to efficiently compare the pre-calculated vectors described at Section~\ref{sec:dataset_enrichment}. We choose one of the queries of the dataset and we use the corresponding embedding of the query to search for semantically similar documents. To achieve this, a cosine similarity comparison is needed between the vector of the query $V_q$ and the vector of a document $V_d$:

\begin{equation}
\cos ({\bf V_q},{\bf V_d})= {{\bf V_q} {\bf V_d} \over \|{\bf V_q}\| \|{\bf V_d}\|} = \frac{ \sum_{i=1}^{n}{{\bf V_q}_i{\bf V_d}_i} }{ \sqrt{\sum_{i=1}^{n}{({\bf V_q}_i)^2}} \sqrt{\sum_{i=1}^{n}{({\bf V_d}_i)^2}} }
\end{equation}

Available approaches:
\begin{itemize}
  \item In the \textbf{naive/brute-force} approach, we perform in-memory all possible comparisons between a given $V_q$ and the $V_d$s of all the documents. This has a complexity of $O(n)$. The final result should be sorted based on the similarity which will involve an additional $O(nlog(n))$, resulting to a total complexity of $O(nlog(n))$.
  \item When using the \textbf{Elasticsearch} framework, we firstly index the documents with their pre-calculated vector embeddings and then we issue the query, using the vector of the query embeddings. We measure the time needed for the indexing and the query answering. We use version 7.17.0 of Elasticsearch which runs as a service in a docker container with one node.
  \item As with the Elasticsearch framework, in the \textbf{FAISS} approach, we index the document embeddings before issuing the query and it takes advantage of this indexing to provide fast answers. We use a Flat index which allows us to perform exact search. The version of FAISS used is 1.5.3 and it runs on CPU (faiss-cpu 1.7.1).
\end{itemize}

The Python implementation of the above ran on a virtual machine with 20GB of RAM and 10-cores of an AMD Ryzen 7 PRO 5850U CPU. We compared the indexing and the querying time for FAISS and Elasticsearch. For the naive/brute-force implementation, we only measured the querying time, since it doesn't perform any pre-processing or indexing. It is worth mentioning that for FAISS and Elasticsearch we asked the top-100 results while the naive/brute-force implementation returned all the results ordered by their similarity search score.

Figure~\ref{fig:naive_faiss_elastic_comparison} shows the results of these experiments for different dataset sizes. All experiments ran three times and we report the average time. On the left sub-figure we demonstrate the time needed for indexing and querying for datasets ranging from 1,000 documents to 80,000 documents. It is worth noting that the y-axis is in logarithmic scale and that the time needed for Elasticsearch to index the document embeddings and to query them is orders of magnitude higher than the other two approaches. Moreover, the largest dataset that we could index with Elasticsearch was 20,000 documents while for the other two approaches we started to have problems for datasets bigger than 80,000 documents. Even in that case, the bottleneck seemed to be the initial loading of the dataset in memory and not necessarily the vector comparison. 

Since Elasticsearch performance is worse than the performance of the other two approaches, on the right sub-figure, we focus on FAISS and the naive/brute-force approach. We observe that FAISS querying is faster than the naive approach. Moreover, if we add the indexing time of FAISS to its querying time, it is still faster than the naive approach. It also scales better than naive when the size of the dataset increases. We expect even better performance if we use the GPU implementation of FAISS.

\subsection{Result quality assessment}
\label{sec:result_quality}
In this section we investigate the quality of the results in two different settings: (a) We verify if the top-k results returned by each vector comparison approach differ and (b) we check if the queries return meaningful results.

To verify (a), we run all approaches against different dataset sizes and we ask for the top-k results of a given query (embedding). We have chosen to return the top-100 results. We compare the returned results from all approaches, one-by-one, and we count how many times any of the approaches returns a different result at a specific position \textit{i} of the top-100 results.
Table~\ref{tab:avg_error} shows the average number of errors found when searching for the top-100 results of a given query for various dataset sizes (from 500 to 10,000 documents). The results were averaged after performing two executions. 

In (b), we evaluate the quality of the embeddings based on the queries asked, the results retrieved and the documents that correspond to these queries. More specifically, we used the first 10,000 documents of the dataset described in Section~\ref{sec:dataset_enrichment} and we asked 100 queries that correspond to the first 100 rows of the dataset. Please note that each line of the dataset contains a document and a query that is related to that document. Then, for every query we verified if the corresponding document of the query is in the top-k of the documents retrieved as answers. The results of this experiments can be found at Table~\ref{tab:correct_res} for different values of \textit{k}.

\begin{table}[tb]
    \begin{minipage}[t]{.47\linewidth}
      \centering
        \begin{tabular}{||c c||} 
         \hline
         Dataset size & Avg. errors \\ [0.5ex] 
         \hline\hline
         500 & 0 \\ 
         \hline
         1,000 & 1 \\
         \hline
         5,000 & 2 \\
         \hline
         10,000 & 1 \\
         \hline
        \end{tabular}
        \caption{\label{tab:avg_error}Average number of differences between FAISS, Elasticsearch and naive/brute-force approaches when retrieving the top-100 results for different dataset sizes.}
    \end{minipage}%
    ~~~~~~~~~~~~
    \begin{minipage}[t]{0.47\textwidth}
      \centering
        \begin{tabular}{||c c||} 
         \hline
         Top-k & Expected results in top-k \\ [0.5ex] 
         \hline\hline
         50 & 16 \\ 
         \hline
         100 & 17 \\
         \hline
         500 & 21 \\
         \hline
         1,000 & 26 \\
         \hline
        \end{tabular}
        \caption{\label{tab:correct_res}Number of expected results that are in the top-k results.}
    \end{minipage} 
\end{table}

\section{Conclusion and future work}
In this paper we have discussed about various word embedding and sentence embedding algorithms. We have selected BERT as our sentence embedding method of preference and we used it to create embeddings for the documents and the questions found at the Google Natural Question dataset. We used the resulting embeddings to evaluate the performance of three vector comparison approaches: FAISS, Elasticsearch and a naive/brute-force approach. The results have shown that FAISS outperforms the other two approaches and that Elasticsearch fails to keep up, especially when bigger datasets are used in a centralized scenario with one node.

As further research, we would like to investigate the performance of Elasticsearch with a cluster of nodes where it can show its benefits against the other two in-memory approaches. Moreover, we would like to examine other efficient vector comparison approaches, like spatial and vector databases. Last but not least, we would like to experiment with different ways of document indexing and query evaluation, like a more fine-grained, paragraph-based, indexing.

\bibliographystyle{unsrt}  
\bibliography{setn2022_sentence_embeddings_comparison}

\begin{thebibliography}{10}

\bibitem{cambria2014icim}
Erik Cambria and Bebo White.
\newblock Jumping nlp curves: A review of natural language processing research
  [review article].
\newblock {\em IEEE Computational Intelligence Magazine}, 9(2):48--57, 2014.

\bibitem{mikolov2013arxiv}
Tomas Mikolov, Kai Chen, Greg Corrado, and Jeffrey Dean.
\newblock Efficient estimation of word representations in vector space, 2013.

\bibitem{mikolov2013nips}
Tomas Mikolov, Ilya Sutskever, Kai Chen, Greg Corrado, and Jeffrey Dean.
\newblock Distributed representations of words and phrases and their
  compositionality.
\newblock In {\em Neural and Information Processing System (NIPS)}, 2013.

\bibitem{weston2011ijcai}
Jason Weston, Samy Bengio, and Nicolas Usunier.
\newblock Wsabie: Scaling up to large vocabulary image annotation.
\newblock In {\em Proceedings of the Twenty-Second International Joint
  Conference on Artificial Intelligence - Volume Volume Three}, IJCAI'11, page
  2764–2770. AAAI Press, 2011.

\bibitem{socher2011icml}
Richard Socher, Cliff Chiung-Yu Lin, Andrew~Y. Ng, and Christopher~D. Manning.
\newblock Parsing natural scenes and natural language with recursive neural
  networks.
\newblock In {\em Proceedings of the 28th International Conference on
  International Conference on Machine Learning}, ICML'11, page 129–136,
  Madison, WI, USA, 2011. Omnipress.

\bibitem{turney2010jair}
Peter~D. Turney and Patrick Pantel.
\newblock From frequency to meaning: Vector space models of semantics.
\newblock {\em J. Artif. Int. Res.}, 37(1):141–188, January 2010.

\bibitem{cambria2017iis}
Erik Cambria, Soujanya Poria, Alexander Gelbukh, and Mike Thelwall.
\newblock Sentiment analysis is a big suitcase.
\newblock {\em IEEE Intelligent Systems}, 32(6):74--80, 2017.

\bibitem{joulin2019arxiv}
Armand Joulin, Edouard Grave, Piotr Bojanowski, Matthijs Douze, Herv{\'{e}}
  J{\'{e}}gou, and Tom{\'{a}}s Mikolov.
\newblock Fasttext.zip: Compressing text classification models.
\newblock {\em CoRR}, abs/1612.03651, 2016.

\bibitem{joachims1998ecml}
Thorsten Joachims.
\newblock Text categorization with support vector machines: Learning with many
  relevant features.
\newblock In Claire N{\'e}dellec and C{\'e}line Rouveirol, editors, {\em
  Machine Learning: ECML-98}, pages 137--142, Berlin, Heidelberg, 1998.
  Springer Berlin Heidelberg.

\bibitem{mccallum1998aaai}
Andrew McCallum and Kamal Nigam.
\newblock A comparison of event models for naive bayes text classification.
\newblock In {\em AAAI 1998}, 1998.

\bibitem{fan2008jmlr}
Rong-En Fan, Kai-Wei Chang, Cho-Jui Hsieh, Xiang-Rui Wang, and Chih-Jen Lin.
\newblock Liblinear: A library for large linear classification.
\newblock {\em J. Mach. Learn. Res.}, 9:1871–1874, jun 2008.

\bibitem{pennington2014emnlp}
Jeffrey Pennington, Richard Socher, and Christopher~D. Manning.
\newblock Glove: Global vectors for word representation.
\newblock In {\em Empirical Methods in Natural Language Processing (EMNLP)},
  pages 1532--1543, 2014.

\bibitem{peters2018arxiv}
Matthew~E. Peters, Mark Neumann, Mohit Iyyer, Matt Gardner, Christopher Clark,
  Kenton Lee, and Luke Zettlemoyer.
\newblock Deep contextualized word representations, 2018.

\bibitem{devlin2019nacacl}
Jacob Devlin, Ming-Wei Chang, Kenton Lee, and Kristina Toutanova.
\newblock {BERT}: Pre-training of deep bidirectional transformers for language
  understanding.
\newblock In {\em Proceedings of the 2019 Conference of the North {A}merican
  Chapter of the Association for Computational Linguistics: Human Language
  Technologies, Volume 1 (Long and Short Papers)}, pages 4171--4186,
  Minneapolis, Minnesota, June 2019. Association for Computational Linguistics.

\bibitem{liu2019arxiv}
Yinhan Liu, Myle Ott, Naman Goyal, Jingfei Du, Mandar Joshi, Danqi Chen, Omer
  Levy, Mike Lewis, Luke Zettlemoyer, and Veselin Stoyanov.
\newblock Roberta: {A} robustly optimized {BERT} pretraining approach.
\newblock {\em CoRR}, abs/1907.11692, 2019.

\bibitem{reimers2019arxiv}
Nils Reimers and Iryna Gurevych.
\newblock Sentence-bert: Sentence embeddings using siamese bert-networks.
\newblock {\em CoRR}, abs/1908.10084, 2019.

\bibitem{yang2019arxiv}
Zhilin Yang, Zihang Dai, Yiming Yang, Jaime~G. Carbonell, Ruslan Salakhutdinov,
  and Quoc~V. Le.
\newblock Xlnet: Generalized autoregressive pretraining for language
  understanding.
\newblock {\em CoRR}, abs/1906.08237, 2019.

\bibitem{lan2019corr}
Zhenzhong Lan, Mingda Chen, Sebastian Goodman, Kevin Gimpel, Piyush Sharma, and
  Radu Soricut.
\newblock {ALBERT:} {A} lite {BERT} for self-supervised learning of language
  representations.
\newblock {\em CoRR}, abs/1909.11942, 2019.

\bibitem{elasticsearch}
Free and open search: the creators of elasticsearch, elk \& kibana, 2022.

\bibitem{kononenko2014msr}
Oleksii Kononenko, Olga Baysal, Reid Holmes, and Michael~W. Godfrey.
\newblock Mining modern repositories with elasticsearch.
\newblock In {\em MSR 2014}, 2014.

\bibitem{lucene}
{Apache Lucene}, 2022.

\bibitem{johnson2021tbg}
Jeff Johnson, Matthijs Douze, and Hervé Jégou.
\newblock Billion-scale similarity search with gpus.
\newblock {\em IEEE Transactions on Big Data}, 7(3):535--547, 2021.

\bibitem{wieschollek2016cvpr}
Patrick Wieschollek, Oliver Wang, Alexander Sorkine-Hornung, and Hendrik P.~A.
  Lensch.
\newblock Efficient large-scale approximate nearest neighbor search on the gpu.
\newblock {\em 2016 IEEE Conference on Computer Vision and Pattern Recognition
  (CVPR)}, pages 2027--2035, 2016.

\bibitem{kwiatkowski2019tacl}
Tom Kwiatkowski, Jennimaria Palomaki, Olivia Redfield, Michael Collins, Ankur
  Parikh, Chris Alberti, Danielle Epstein, Illia Polosukhin, Matthew Kelcey,
  Jacob Devlin, Kenton Lee, Kristina~N. Toutanova, Llion Jones, Ming-Wei Chang,
  Andrew Dai, Jakob Uszkoreit, Quoc Le, and Slav Petrov.
\newblock Natural questions: a benchmark for question answering research.
\newblock {\em Transactions of the Association of Computational Linguistics},
  2019.

\end{thebibliography}

\end{document}